%% file: main.tex
\documentclass[11pt,a4paper]{article}
\usepackage[utf8]{inputenc}
\usepackage[hyperref]{acl2021}
\usepackage{times}
\usepackage{latexsym}

\usepackage{microtype}

\usepackage{linguex}
\AtBeginDocument{\settowidth{\Exlabelwidth}{(0)}}

\usepackage{multicol} 
\usepackage{verbatim}
\usepackage{import}
\usepackage{graphicx}
\usepackage{enumitem}
\usepackage{booktabs}
\usepackage{xspace}
\usepackage[colorinlistoftodos]{todonotes}

\usepackage{enumitem}
\usepackage{fdsymbol}

\aclfinalcopy 
\def\aclpaperid{4070} 

\renewcommand{\d}[1]{\textcolor{blue}{DH: #1}}

\newcommand{\fullName}[0]{\textsf{Full}\xspace}
\newcommand{\noNPIName}[0]{\textsf{Full$\setminus$NPI}\xspace}
\newcommand{\noENVName}[0]{\textsf{Full$\setminus$ENV$\cap$NPI}\xspace}

\title{Language Models Use Monotonicity to Assess NPI Licensing}

\author{Jaap Jumelet$^\uparrow$~~~~
  Milica Denić$^\uparrow$~~~~
  Jakub Szymanik$^\uparrow$\\
  \textbf{Dieuwke Hupkes}$^\lambda$~~~~
  \textbf{Shane Steinert-Threlkeld}$^\downarrow$\\[5pt]
  $^\uparrow$ Institute for Logic, Language and Computation, University of Amsterdam\\
  $^\lambda$ Facebook AI Research\\
  $^\downarrow$ Department of Linguistics, University of Washington\\[5pt]
  \small{\texttt{\{j.w.d.jumelet, m.denic, J.K.Szymanik\}@uva.nl}~~~~~~\texttt{dieuwkehupkes@fb.com}~~~~~~\texttt{shanest@uw.edu}}
  }

\begin{document}

\maketitle

\input{sections/abstract}

\input{sections/introduction}

\input{sections/background}

\input{warstadt_corpus}

\input{sections/related}

\input{sections/methods}

\input{sections/experiments}

\input{sections/conclusion}

\section*{Acknowledgments}
We thank Oskar van der Wal and Lucas Weber for their valuable feedback.
We thank Jack Hoeksema for providing us with the list of NPIs.
MD and JS were funded by the European Research Council under the European Unions Seventh Framework Programme (FP/20072013)/ERC Grant Agreement n. STG 716230 CoSaQ.

\bibliography{anthology,acl2020}
\bibliographystyle{acl_natbib}

\newpage
\input{sections/appendix-npis.tex}

\end{document}

%% file: sections/abstract.tex
\begin{abstract}
We investigate the semantic knowledge of language models (LMs), focusing on (1) whether these LMs create categories of linguistic {environments} based on their semantic \emph{monotonicity} properties, and (2) whether these categories play a similar role in LMs as in human language understanding, using negative polarity item licensing as a case study. 
We introduce a series of experiments consisting of probing with diagnostic classifiers (DCs), linguistic acceptability tasks, as well as a novel \textit{DC ranking} method that tightly connects the probing results to the inner workings of the LM.
By applying our experimental pipeline to LMs trained on various filtered corpora, we are able to gain stronger insights into the semantic generalizations that are acquired by these models.\footnote{All code and data can be found at \url{https://github.com/jumelet/monotonicity-npi-lm}}
\end{abstract}

%% file: sections/introduction.tex
\section{Introduction}

Neural language models (LMs) have become powerful approximators of human language, making it increasingly important to understand the features and mechanisms underlying their behavior \citep{blackbox18,blackbox19}.
In the past few years, a substantial number of studies have investigated the linguistic capabilities of LMs \citep[i.a.]{gulordava-etal-2018-colorless,giulianelli-etal-2018-hood,lakretz-etal-2019-emergence,DBLP:journals/corr/abs-2012-05395,ettinger2020bert}.
Such work has focused primarily on \emph{syntactic} properties, while fewer studies have been done on what kind of \emph{formal semantic} features are encoded by language models.
In this paper, we focus explicitly on what LMs learn about a semantic property of sentences, and in what ways their knowledge reflects well-known features of human language processing. 

As the topic of our studies, we consider \textbf{monotonicity}, a semantic property of linguistic environments that plays an important role in human language understanding and inference \citep{hoeksema1986monotonicity, valencia1991studies, vanbenthem1995language, Icard-III:2014xy}: the monotonicity of a linguistic environment determines whether inferences from a general to a particular term or vice versa are valid in that environment.  
For example, the fact that the inference from ``\textit{Mary didn't write a paper}'' to ``\textit{Mary didn't write a linguistics paper}'' is valid shows us that the position where ``\textit{a paper}'' occurs is \emph{downward monotone}: the inference is valid when a more general term (``\textit{a paper}'') is replaced with a more specific one (``\textit{a linguistics paper}'').

To investigate monotonicity we focus on \textbf{negative polarity items} (NPIs): a class of expressions such as \textit{any} or \textit{ever} that are solely acceptable in downward monotone environments \cite{Fauconnier, LadusawDiss}.
Psycholinguistic research has confirmed this connection between NPIs and monotonicity: humans judge NPIs acceptable in a linguistic environment if they consider that environment to be downward monotone \citep{Chemla:2011aa}.
Previous research has established that LMs are relatively successful in processing NPIs \citep{warstadt-etal-2019-investigating}, but without investigating \emph{how} they came to these successes. 

We raise the following research questions: \begin{itemize}\setlength\itemsep{0mm}\vspace{-2.0mm}
    \item[\textbf{RQ1}] Do language models encode the monotonicity properties of  linguistic environments?
    \item[\textbf{RQ2}] To what extent do they employ this information when processing negative polarity items?
\end{itemize}


We developed a series of experiments, in which we first evaluate the general capacities of LMs in handling monotonicity and NPIs and then investigate the generalization heuristics of the LM by doing experiments with modified training corpora.
First, we establish that LMs are able to encode a notion of monotonicity by probing them with diagnostic classifiers \citep[DCs,][]{Hupkes2018} (\S\ref{sec:exp1}).
In our second experiment we demonstrate that our LMs are reasonably successful with NPI licensing using an NPI acceptability task (\S\ref{sec:exp2}). 
Next, we introduce a novel \emph{DC ranking} method to investigate the overlap between the information that the model uses to make judgments about NPIs and the information that the DCs use to predict monotonicity information, finding that there is a significant overlap (\S\ref{sec:exp3}). 

We then investigate two potential confounds that may obfuscate our results.
First, we consider whether the signal that is picked up by the monotonicity DC is not simply a proxy that tells the model that an NPI may occur at that position (\S\ref{sec:exp4}).
To assess this, we train new LMs on a corpus from which all sentences with NPIs have been removed, re-run the montonicity probing task, and find that even in the absence of NPI information, LMs are still able to encode a notion of monotonicity.

Next, we consider whether an LM bases its NPI predictions on simple co-occurrence heuristics, or if it can extrapolate from a general notion of monotonicity to cases of NPIs in environments in which they have never been encountered during training (\S\ref{sec:exp5}).
We again train new LMs on modified corpora, this time removing NPIs only in one specific environment, and repeat the NPI acceptability and DC ranking experiments.
The results of this setup demonstrate that LMs indeed use a general notion of monotonicity to predict NPI licensing.

\paragraph{Contributions}
With this work, we contribute to the ongoing study of the linguistic abilities of language models in several ways:
\begin{itemize}\setlength\itemsep{0mm}\vspace{-1.0mm}
    \item With a series of experiments we demonstrate that LMs are able to acquire a general notion of monotonicity that is employed for NPI licensing.
    \item We present two novel experimental setups: \emph{filtered corpus training} and \emph{DC ranking}, that can be used to assess the impact of specific information during training and compare the information used by DCs with the information used with the model, respectively.
    \item By using experimental results from psycho-semantics to motivate hypotheses for  LM behavior, we find that our models reflect behavior similar to human language processing.
\end{itemize}

In the remainder of this paper, we will first provide some linguistic background that helps to situate and motivate our experiments and results (\S\ref{sec:background}).
We then discuss related work on NPI processing in LMs in \S\ref{sec:related}.
In \S\ref{sec:methods}, we discuss our methods and experimental setup.
\S\ref{sec:exp1} through \S\ref{sec:exp5} explain and present the results.
We conclude in \S\ref{sec:discussion} with a general discussion and pointers to future work.

%% file: sections/background.tex
\section{Linguistic Background}\label{sec:background}


\paragraph{Monotonicity}

Monotonicity is a property of a linguistic environment which determines what kind of inferences relating general and particular terms are valid in that environment. 
If inferences from a general to a particular term are valid, the linguistic environment is said to be \textit{downward monotone} (DM). 
If inferences are valid the other way around, from a particular to a general term, the linguistic environment is said to be \mbox{\textit{upward monotone} (UM)}. 

Examples of expressions inducing DM environments are negation and quantifiers like \emph{nobody, no NP}, but also specific types of adverbs and the antecedents of conditional sentences. 
For instance, \Next below exemplifies that in these environments the inference from a sentence with a general term (\emph{cookies}) to that sentence with a more particular term (\emph{chocolate cookies}) is valid, but not vice versa.

\ex. 
\a. Mary did\textbf{n't} eat cookies. $\Rightarrow$\\Mary didn't eat \textit{chocolate} cookies. \label{de-inf}
\b. \textbf{Nobody} ate cookies. $\Rightarrow$\\Nobody ate \textit{chocolate} cookies.
\c. Mary \textbf{rarely} ate cookies. $\Rightarrow$\\Mary rarely ate \textit{chocolate} cookies.

Common examples of UM environments are (non-quantified) positive sentences, quantifiers such as \emph{somebody, many NP}, and other kind of adverbs. 
\Next exemplifies that in these environments the inference from a sentence with a more particular term (\emph{chocolate cookies}) to the same sentence with a general term (\emph{cookies}) is valid, but not vice versa.

\ex.
\a. Mary ate \textit{chocolate} cookies.  $\Rightarrow$\\ Mary ate cookies. \label{ue-inf}
\b. \textbf{Everyone} ate \textit{chocolate} cookies. $\Rightarrow$ Everyone ate cookies.
\c. Mary \textbf{often} ate \textit{chocolate} cookies. $\Rightarrow$ Mary often ate cookies.



\paragraph{NPIs}
NPIs are expressions such as the English words \textit{any, anyone, ever}, whose acceptability depends on whether its linguistic environment is downward monotone \citep{Fauconnier, LadusawDiss, dowty1994role, kadmon1993, krifka1995, lahiri98, chierchia2006broaden, chierchia2013}.\footnote{See however \citealp{zwarts1995nonveridical, giannakidou1998polarity, barker2018negative} for different takes on NPI acceptability generalizations.} 
While the conditions for NPI acceptability are complex, a good approximation is that NPIs are acceptable (or \textit{licensed}) in the syntactic scope of \textit{NPI licensors} that induce a DM environment.\footnote{An NPI occurs in the syntactic scope of a licensor if the licensor \textit{c-commands} the NPI.
An NPI licensor {c-commands} an NPI if the NPI is the licensor's sister node or one of its sister's descendants in a constituent tree \citep{reinhart1976syntactic}.
}
If we again consider the DM environment of \ref{de-inf} and the UM environment of \ref{ue-inf}, it can be seen that English \emph{any} is an NPI, as it is acceptable when inside the syntactic scope of negation (a DM expression) as in \Next[a], and not acceptable when they are in an UM environment as in \Next[b].

\ex.
\a. Mary didn't eat (any) cookies. \label{no-cookies}
\b. Mary ate (*any) cookies.

Importantly, monotonicity plays a role at the psychological level: human judgments about the monotonicity of a linguistic environment predict their judgments of NPI acceptability in that environment \citep{Chemla:2011aa, Denic-PIinferences}. 
For example, how plausible someone finds the inference \ref{de-inf} predicts how acceptable they find the sentence \ref{no-cookies}.
Summing up, NPI licensing has a syntactic component (NPIs must reside in syntactic scope of a licensor) and a semantic component (NPI licensors are DM expressions), that are connected on a psychological level (monotonicity judgments predict NPI acceptability).
Our research aims to uncover whether this connection is exhibited by LMs as well.



%% file: warstadt_corpus.tex
\begin{table*}[ht]
\scriptsize
\centering
\begin{tabular}{l l l l}
    \toprule
    Environment Class & Abbrev. & DM example & UM example\\\midrule
    Adverbs & \textsf{ADV} & A lady \textbf{rarely} \textit{ever} ... & *A lady sometimes \textit{ever}...\\
    Conditionals & \textsf{COND} & \textbf{If} the dancers see \textit{any} ... & *While the dancers see \textit{any}...\\
    Determiner Negation & \textsf{D-NEG} & \textbf{No} teacher says that the students had practiced \textit{at all}. & *Some teacher says that the students had practiced \textit{at all}.\\
    Sentential Negation & \textsf{S-NEG} & The dancer was \textbf{not} saying that the guy had profited \textit{yet}. & *The dancer was really saying that the guy had profited \textit{yet}.\\
    Only & \textsf{ONLY} & \textbf{Only} the boys had \textit{ever} ... & *Even the boys had \textit{ever} ...\\
    Quantifiers & \textsf{QNT} & \textbf{Every} senator who had \textit{ever} ... & *Some senator who had \textit{ever} ...\\
    Embedded Questions & \textsf{QUES} & The patients wonder \textbf{whether} the lady admires \textit{any} ... & *The patients say that the lady admires \textit{any}...\\
    Simple Questions & \textsf{SMP-Q} & Did the boy \textit{ever} listen\textbf{?} & *The boy did ever listen.\\
    Superlatives & \textsf{SUP} & A lady buys the \textbf{oldest} dish that the adult had \textit{ever} ... & *A lady buys the old dish that the adult had \textit{ever} ... \\\bottomrule
\end{tabular}
\caption{The nine environment classes of \citet{warstadt-etal-2019-investigating}, with an example of a minimal DM/UM pair for each class taken from the corpus.}
\label{tab:warstadt-overview}
\end{table*}

%% file: sections/related.tex
\section{Related work}\label{sec:related}
The literature on interpreting LMs has grown substantially in the last few years \citep[see, e.g.][for survey papers]{Belinkov2019, Alishahi2019, rogers2021primer}.
Several studies investigate how they process NPIs, focused mainly on the \emph{syntactic} aspect of NPI licensing. 

\citet{Jumelet2018} conclude that LSTM language models encode information about the dependency between the NPI and the NPI licensor, although this effect diminishes as the distance between the NPI and its licensor grows.
\citet{Marvin2019} study NPI judgments of LMs on minimally different sentence pairs (with the NPI licensor either in an appropriate syntactic configuration or not) and find that their models are unable to reliably assign higher probability to sentences in which NPIs are correctly licensed. 
The syntactic aspect of NPI licensing is also examined by \citet{futrell-etal-2019-neural}, who demonstrate that LSTM LMs are susceptible to learning spurious licensing relationships, a finding that \citet{DBLP:warstadt2020} demonstrate to also hold for BERT \citep{DBLP:conf/naacl/DevlinCLT19}.
\citet{Wilcox2019} investigate how explicit syntactic supervision of LMs affects their success with syntactic aspects of NPI licensing.
The broad linguistic suites of \citet{warstadt2019blimp} and \citet{hu-etal-2020-systematic} also contain a set of tasks related to NPI licensing, demonstrating that it is one of the most challenging tasks for LMs to handle.
\citet{DBLP:journals/corr/abs-2101-11287} investigated the dynamics of NPI learning during training, and connected this to a multi-task learning paradigm, demonstrating that LMs are able to efficiently leverage information from related licensing environments.

Lastly, \citet{warstadt-etal-2019-investigating} examine BERT's ability in determining NPI acceptability. 
They demonstrate that BERT has significant knowledge of the dependency between NPIs and their licensors, but that this success varies widely across different experimental methods. 
Our study builds on that of \citet{warstadt-etal-2019-investigating}. 
Although they demonstrate that BERT is generally successful with NPI licensing, their results do not reveal whether BERT has constructed a more general category of DM expressions that is independent of collocational cues, nor whether it has understood that this category matters for NPI licensing.

%% file: sections/methods.tex
\section{Methods}\label{sec:methods}

Before getting to the main experimental part of our work, we briefly discuss the training corpus, model architecture and evaluation corpus we consider.

\paragraph{Training Corpus}
The base training corpus we consider in our experiments is the corpus used by \citet{gulordava-etal-2018-colorless}.
This corpus is a collection of sentences from Good and Featured English Wikipedia articles and consists of over 90M tokens. 
The vocabulary of the corpus consists of the 50.000 most frequent tokens in this corpus; less frequent tokens are mapped to a special \texttt{<unk>} token.
We refer to the full training corpus type with the name \fullName, and to the LMs trained on this corpus as \fullName~LMs. 
In addition to \fullName, we use multiple other corpora which are derived from \fullName~by means of filtering. 
This will allow us to draw conclusions about specific generalization abilities and reliance on collocational cues of LMs; filtered corpora will be introduced in the relevant sections.

\paragraph{Model Architecture}
In our studies, we focus on recurrent language models.
More specifically, following \citet{gulordava-etal-2018-colorless}, we consider two-layer LSTM language models, with an embedding and hidden size of 650.
All training runs across our experiments follow the same regime, identical to the regime described by \citet{gulordava-etal-2018-colorless}: 40 epochs of training with SGD, with a plateau scheduler and an initial learning rate of 20, a batch size of 64, BPTT length of 35, and dropout of 0.1.\footnote{Models are trained on a \textit{GeForce 1080 Ti} GPU, take around 40 hours to train, and consist of ~71M parameters.}

\paragraph{Evaluation Corpus}
To assess monotonicity and NPI licensing knowledge of LMs in our experiments, we leverage the NPI corpus of \citet{warstadt-etal-2019-investigating}, which consists of a large amount of grammatical and ungrammatical sentences with NPIs.
This corpus is divided into 9 distinct \textbf{environment classes}, allowing for fine-grained analysis of NPI licensing.
Importantly, these nine environment classes come in two versions: a DM version---in which NPIs are grammatically acceptable, and a minimally different UM version---in which they are not.
We provide an overview with examples of DM and UM versions of all environment classes in Table~\ref{tab:warstadt-overview}.
The full size of the corpus is 106.000 distinct DM sentences, and the division of environment classes is split roughly uniformly.

%% file: sections/experiments.tex
\begin{figure*}
    \centering
    \includegraphics[width=\textwidth]{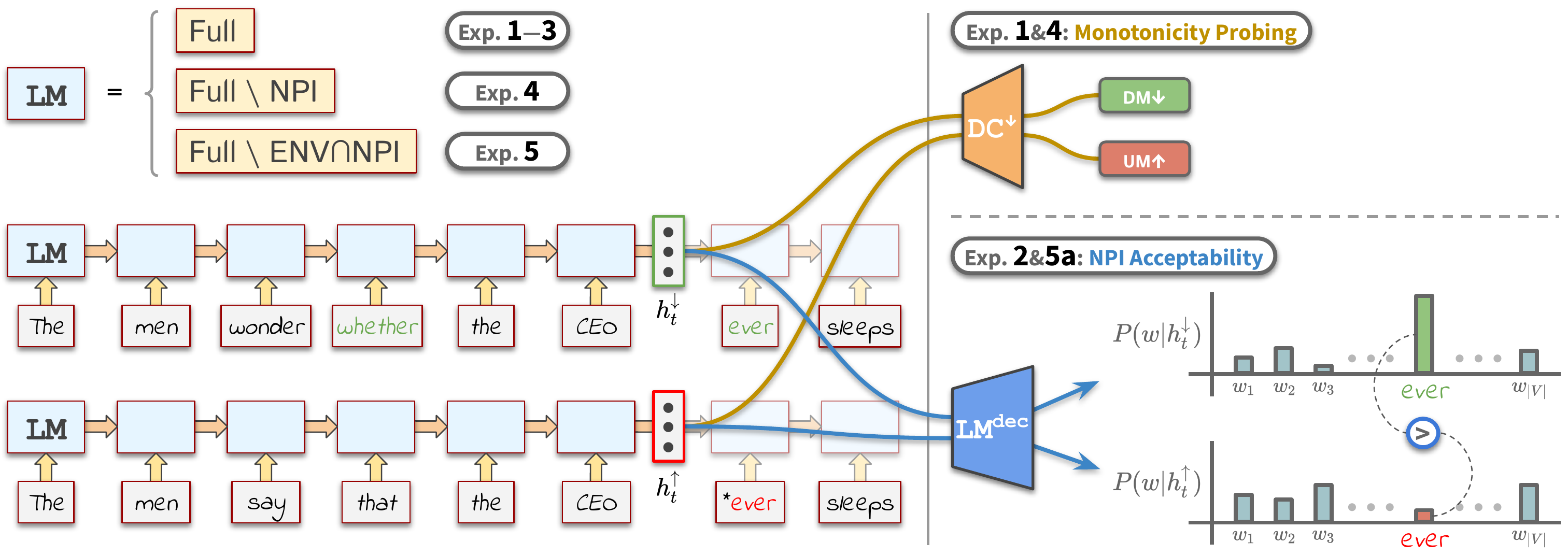}
    \caption{
    The pipeline of our experimental setup.
    We start by computing the hidden states $h^\downarrow_t$ (within a DM environment ahead of the NPI) and $h^\uparrow_t$ (within a UM environment).
    These hidden states are then used for training the monotonicity DC (Exp. 1 \& 4), and to compare $P_\textsc{lm}(\textsc{npi}|h^\downarrow_t) > P_\textsc{lm}(\textsc{npi}|h^\uparrow_t)$ (Exp. 2 \& 5a).
    The task of Experiments 3 and 5a can be found in Figure~\ref{fig:dc_ranking_diagram}.
    Experiments 4 and 5 consist of the same tasks as the first three experiments, but differ in the language model that is used.
    }
    \label{fig:experimental_overview}
\end{figure*}

\section{Experiments and Results}\label{sec:experiments}
In this section we describe the experimental pipeline in more detail.
A graphical overview of our experiments is depicted in Figures~\ref{fig:experimental_overview} and \ref{fig:dc_ranking_diagram}.
Each experiment description is directly followed by an \mbox{analysis} of its results. 

\subsection{Experiment 1: Do LMs represent monotonicity information?} \label{sec:exp1}
In our first experiment, we test whether LMs trained on our \fullName~corpus possess a notion of monotonicity.
We train five different LMs and test how well they represent monotonicity properties of different environments by training linear \textbf{diagnostic classifiers} \citep[DCs,][]{Hupkes2018} on top of the hidden states of the LM.
To create a corpus of monotonicity sentences for training and testing the DCs, we leverage the corpus of \citet{warstadt-etal-2019-investigating}, now selecting all DM and UM sentences to build up a balanced corpus of these categories.
The nine environment classes in that corpus hence provide a broad spectrum of DM environments and their minimally different UM counterparts. 

For training and testing the DCs, we consider the hidden states at the position directly before an NPI occurs (see Figure~\ref{fig:experimental_overview}).
The reason we train the DCs at this position is because only at this point we are sure that the monotonicity information should surface and be encoded \emph{linearly}.
This is due to the fact that the decoder of the LM that transforms a hidden state into a probability distribution is linear as well: if the probability of some token depends on a linguistic feature, this feature must hence be encoded linearly.
The DCs are implemented using the \texttt{diagNNose} library of \citet{jumelet-2020-diagnnose}, and trained using 10-fold cross-validation, Adam optimization \citep{DBLP:journals/corr/KingmaB14}, a learning rate of $10^{-2}$ and L1 regularization with $\lambda = 0.005$.

We train our monotonicity DCs in two separate ways.
First, we divide the entire monotonicity corpus into a 90/10 train/test split, sampled \textit{uniformly} across the different environment classes.
This allows us to examine whether DM and UM environments are linearly separable in a way that is applicable to all environment classes.
We refer to this classifier as the \textsf{All-ENV} DC.

Second, we move to a more fine-grained type of analysis.
High performance of the \textsf{All-ENV} DC namely does not provide evidence that monotonicity is encoded the same way for each environment: the set of salient hidden units used by the \textsf{All-ENV} DC for classifying monotonicity within the \textit{Adverbs} environment, for example, could be disjoint from the set of units used for the \textit{Only} environment.
To investigate this, we train a DC on the hidden states of all-but-one environment class, and test its performance on the excluded class.
This provides a measure to what extent the monotonicity representation of DM and UM environments derived from all other environment classes \textit{generalizes} to the held-out class, demonstrating stronger evidence that the model represent monotonicity in the same way across different environments.

\begin{figure*}
    \centering
    \includegraphics[width=\textwidth,clip,trim={0 4cm 0 3cm}]{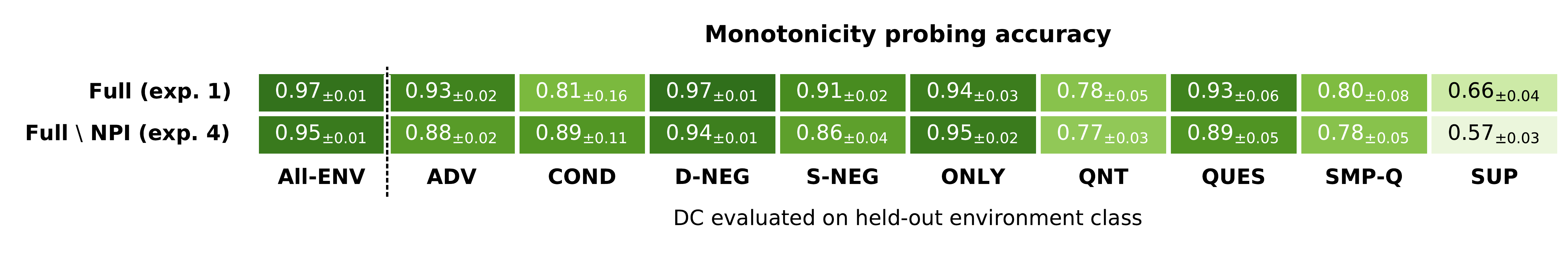}
    \caption{
    Accuracy and standard deviation on the monotonicity diagnostic classification task, averaged over 5 seeds for each model type. 
    The \textsf{All-ENV} column denotes train/test split procedure sampled uniformly over all environment class; other columns denote accuracy on one environment class that has been excluded during training. 
    }\label{fig:exp1}
\end{figure*}
\begin{figure*}[ht]
    \centering
	\includegraphics[width=\textwidth,clip,trim={0 4cm 0 3cm}]{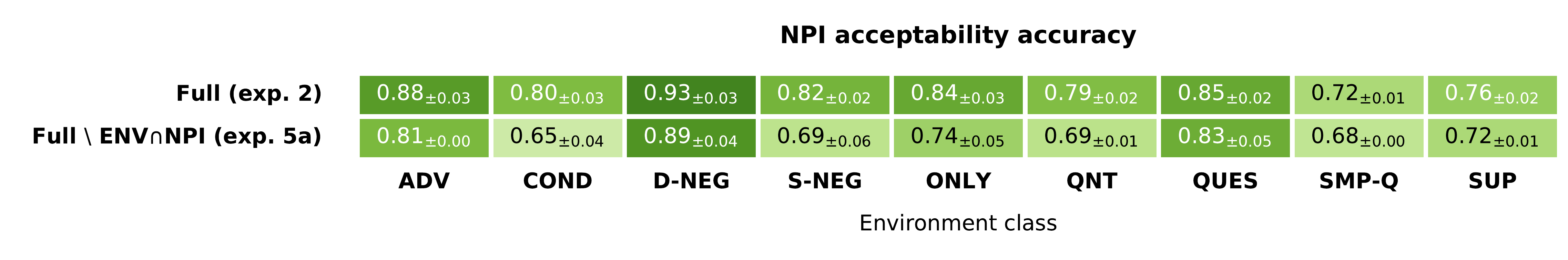}
	\vspace{-0.5cm}
	\setlength{\belowcaptionskip}{-10pt}
	\caption{
	Accuracy on the NPI acceptability task---based on whether the NPI was assigned a higher probability in the DM environment than in its UM counterpart.
	}\label{fig:npi_acceptability}\vspace{1cm}
\end{figure*}

\paragraph{Results}
The results of our first experiment are shown in the top row of Figure~\ref{fig:exp1}.
The first column contains the average accuracy for the \textsf{All-ENV} DC, and it can be seen that the diagnostic classifier succeeds in this task with high accuracy (97\%).
This indicates that the uniform split over all environment classes is linearly separable.

Next, we consider the held-out evaluation procedure for each of the nine environment classes.
It can be seen that the monotonicity signal generalizes well to five classes (\textit{adverbs, determiner negation, only, sentential negation, and embedded questions}), all with an accuracy above 90\%.
The other four classes yield a higher standard deviation, indicating that these classes are encoded less consistently across initialization seeds.
The accuracy for all held-out DCs is lower compared to the \textsf{All-ENV} DC results, indicating that the \textsf{All-ENV} DC relied partly on information unrelated to a shared notion of monotonicity.
The fact that the accuracy of these DCs is still so high, however, indicates that there is a substantial overlap between the way that monotonicity is encoded within the different environments.

\subsection{Experiment 2: Do LMs predict the licensing conditions of NPIs?}\label{sec:exp2}
In the next experiment we investigate the NPI acceptability judgments of the \fullName~LMs on the corpus of \citet{warstadt-etal-2019-investigating}.
This is done by comparing the probability of an NPI conditioned on the model's representation of a DM environment ($h^\downarrow_t$) and a UM environment ($h^\uparrow_t$), where success is defined as follows:
\[P_{\textsc{lm}}(\textsc{npi}|h^\downarrow_t) > P_{\textsc{lm}}(\textsc{npi}|h^\uparrow_t)\]
This is a common evaluation procedure in the interpretability literature \citep{Linzen2016}, and has earlier been applied in the domain of NPI licensing by \citet{Jumelet2018} and \citet{warstadt2019blimp}.
Our approach is similar to the Cloze Test of \citet{warstadt-etal-2019-investigating}, but their setup used (bi-directional) masked LMs, making it possible to directly compare the probabilities of the NPI licensor, instead of comparing the NPI probabilities.
Note that we purposefully do not base NPI acceptability on comparing full sentence probabilities: in our view this type of comparison can be distracted by token probabilities not related to the NPI itself.

We split this procedure out for each of the nine environment classes.
The example sentence of the Simple Questions environment in Table~\ref{tab:warstadt-overview}, for example, is evaluated as follows:
\[P_{\textsc{lm}}(\textit{ever}|\textit{Did the boy}) > P_{\textsc{lm}}(\textit{ever}|\textit{The boy did})\]
Using the full sentence probabilities for this comparison would require taking probabilities into account such as $P_{\textsc{lm}}(the|Did)$ and $P_{\textsc{lm}}(boy|The)$, that have no relation to NPI licensing at all.

\paragraph{Results}
We present the results for this experiment in the top row of Figure~\ref{fig:npi_acceptability}.
The \fullName models demonstrate a considerable ability at predicting NPI acceptability, with the least performing class (\mbox{\textsf{SMP-Q}}, \textit{Simple Questions}) yielding an accuracy that is still well above chance (0.72).
Compared to earlier investigations on the ability of LSTM LMs in NPI licensing, our results indicate that these models are able to obtain a more sophisticated understanding of NPIs than previously thought: both \citet{Marvin2019} and \citet{hu-etal-2020-systematic} report LSTM performance below chance on NPI acceptability tasks.
This might in part be due to the different evaluation procedure we used (conditional vs. full-sentence probability comparison).

\subsection{Experiment 3: Is the LM's knowledge of DM environments and of NPI licensing related?}\label{sec:exp3}
We have now established that our models encode a signal related to monotonicity, and are successful at predicting NPI acceptability.
In our third experiment, we assess to what extent the parameters used by the LM to predict NPIs (i.e.\ the LM's decoder embeddings for NPIs) \emph{overlap} with the information the DCs use to predict the monotonicity properties of a particular environment class.
For this we have devised a novel \textbf{DC ranking} method, that ranks the LM's decoder weights for all tokens based on their similarity with the DC weights.

\begin{figure}
    \centering
    \includegraphics[width=\columnwidth]{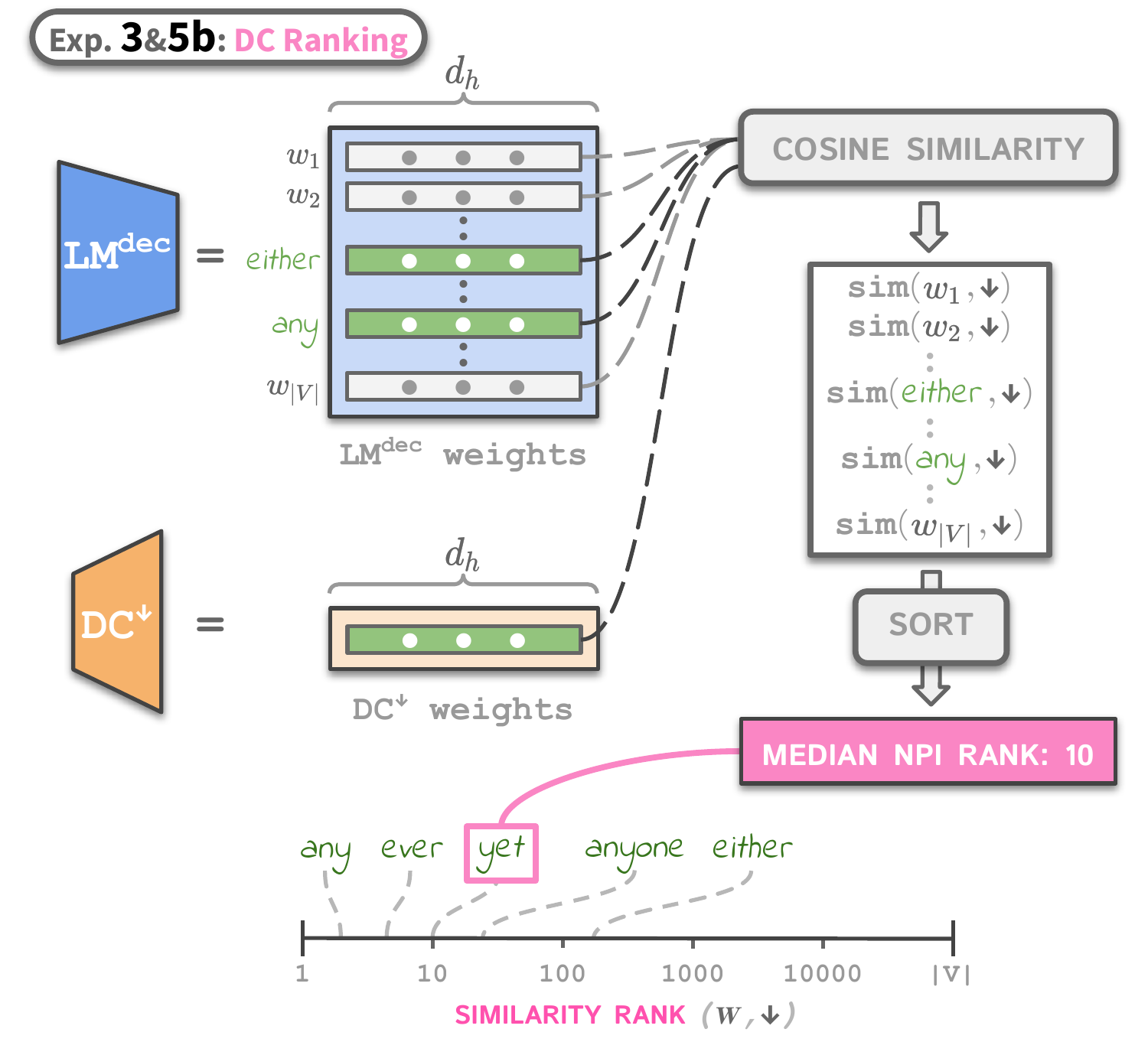}
    \caption{
    The \textit{DC Ranking} experiment, in which we investigate whether the monotonicity DC and the LM decoder base their predictions on similar cues, by computing and ranking the cosine similarities between the DC weights and the decoder weights of each token.
    }
    \label{fig:dc_ranking_diagram}
\end{figure}

\begin{figure*}[ht]
    \centering
	\includegraphics[width=\textwidth,clip,trim={0 4cm 0 3cm}]{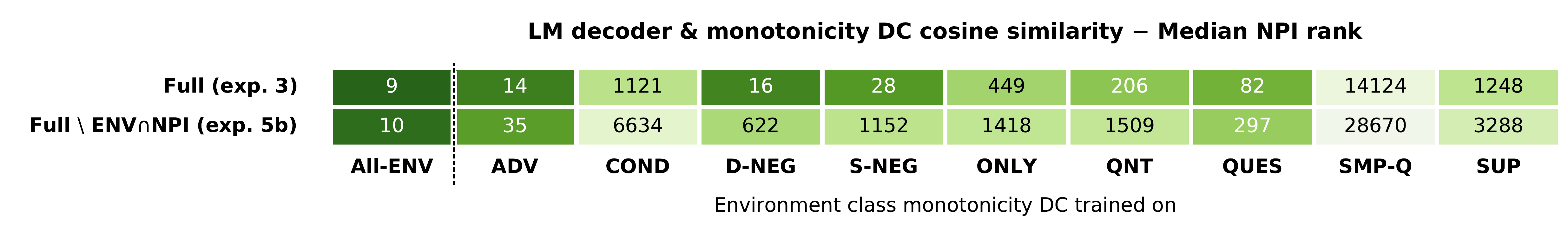}
	\caption{Results on the median NPI rank task.
	A low median rank indicates that the monotonicity DC uses the same representational information as the NPI decoder.
	}\label{fig:npi_rank}
\end{figure*}

We present a schematic overview of the method in Figure~\ref{fig:dc_ranking_diagram}.
The LM's decoder weight matrix can be interpreted as a collection of vectors corresponding to each token in the model's vocabulary.
The monotonicity DC is a binary classifier, so its weights are represented by a single vector.
The LM's decoder vectors are of the same dimensionality as the weight vector of the monotonicity DC, which allows us to compute the similarity between each decoder vector and the monotonicity DC.
For each of the 50.000 tokens in the LM's vocabulary, we calculate the cosine similarity between the decoder weights corresponding to that token and the DC's weights.
We then sort these similarity scores, which results in a ranking of tokens that are most similar to the DC.  

As we are interested in finding the connection of the monotonicity DC and the LM's NPI processing in general, we compute the \textbf{median rank} over a set of 11 NPIs.\footnote{We consider the following 11 single-token NPI expressions: \textit{dared, any, anybody, anymore, anyone, anything, anywhere, ever, nor, whatsoever,} and \textit{yet}. 
These are all the single-token NPIs taken from a list of NPIs that is described in \S\ref{sec:exp4}.}
A low median NPI rank indicates that the LM uses the same cues for NPI prediction as the monotonicity DC, demonstrating a clear connection between NPI licensing and monotonicity. 

Contrary to Experiment~1, we no longer make use of the hold-one-out training procedure, that gave insights to what extent a general monotonicity signal generalizes to a held-out environment class.
Instead, we train a separate diagnostic classifier for each environment class using a train/test split made up of DM and UM environments within that class.
This results in a classifier that represents the class-specific decision boundary between minimal pairs of DM and UM items and allows us to investigate to what extent these decision boundaries align with the weights of the LM decoder.
Next to the environment-specific DCs we also report the DC ranking outcome for the \textsf{All-ENV} DC that has been trained on all environments.

\paragraph{Results}
The results of this experiment are presented in the top row of Figure~\ref{fig:npi_rank}.
The first column (\textsf{All-ENV}) contains the result for the DC trained on all environment classes, and the median NPI rank of 9 demonstrates that the monotonicity DC aligns very closely with the NPI decoder weights of the LMs.
This median rank should be interpreted within the context of the model vocabulary size: it can range upwards to 50.000, so a rank that is close to 0 signifies a tight connection between the probing task and the tokens of interest.

Moving on to the environment-specific results, it can be seen that the results vary considerably between the environment classes.
The worst scoring class is again that of Simple Questions.
This makes sense, as the licensing conditions for question constructions do not depend on the presence of a specific licensing token such as \textit{not}, but on the overall structure of the whole sentence.
The other environment classes lead to scores far closer to 0, indicating that for these classes monotonicity classification is closely aligned to NPI processing.

Interestingly, the median rank of the \textsf{All-ENV} DC is lower than the ranks of all other DCs.
This shows that the model has aligned its representation of NPIs to an aggregate of the monotonicity representations in the different environment classes.
This allows the model to flexibly deal with NPIs in a wide range of licensing environments.

\subsection{Experiment 4: Are NPIs important for learning monotonicity information?}\label{sec:exp4}
With Experiment 3 we established that NPI processing and monotonicity are related in our LMs. 
Now, we investigate to what extent their representations are entangled during training.
More specifically, we investigate if the signal from the presence of NPIs is indispensable for the LM to develop a notion of monotonicity, or if instead the success in categorizing monotonicity environments can be learned independently of NPIs.

We address this question by testing whether LMs can still classify the monotonicity properties of environments when they are completely deprived of NPIs during training.
To do so, we train new language models on a modified corpus that does not contain any NPIs at all.
To arrive at this corpus, we remove all sentences that contain at least one NPI expression from the \fullName corpus.
We identify these expressions based on a comprehensive list of NPI expressions in English collected by \citet{hoeksema2012natural} and the list of NPIs in English compiled by \citet{israel2011grammar}. 
From this list, we manually removed expressions that have both NPI and non-NPI uses (e.g.\ \textit{a thing, a bit}). 
The 40 NPI expressions that resulted from this procedure can be found in Appendix~\ref{appendix-npis}.
We train 5 models on this corpus and refer to them by the name \noNPIName.

In this experiment, we run the monotonicity probing procedure of Experiment~1 on the \noNPIName models.
We posit that if the notion of monotonicity can be learned independently of NPIs, there should be no significant drop in performance compared to the results of the \fullName LMs.

\paragraph{Results}
We report the results of this experiment in the bottom row of Figure~\ref{fig:exp1}.
Again it can be noted that the \textsf{All-ENV} DC, trained and tested uniformly over all environment classes, obtains a high accuracy on the task (0.95).
Furthermore, none of the held-out environment DCs lead to significant drops in performance compared to the \fullName LMs.
Based on this we conclude that even in the absence of NPI cues, LMs are still able to build up a shared robust notion of monotonicity.

\subsection{Experiment 5: How robust is the connection between monotonicity and NPI processing?}
\label{sec:exp5}
This research aims to uncover whether LMs possess a robust connection between monotonicity and NPI licensing.
Our findings indicate that this connection is present in our models. 
A major confound that has not yet been addressed, however, is the extent to which our models rely on collocational cues when judging the acceptability of an NPI.
To test this, we examine whether an LM's connection between NPIs and monotonicity \textit{generalizes} to novel environment classes in which NPIs have never been encountered during the training phase of the LM. 

We have created nine modified corpora in which sentences with NPIs within a specific environment have been removed.
For these different corpora, we again consider the nine NPI-licensing environments of \citet{warstadt-etal-2019-investigating}.
For each environment class we create a new corpus by removing all sentences from the \fullName corpus in which an NPI expression from Appendix~\ref{appendix-npis} is preceded by an expression belonging to that class, somewhere earlier in the sentence.\footnote{For Simple Questions we remove a sentence if an NPI occurs in a sentence that ends with a question mark.} 
Note that we only remove the sentences in which the environment actually licenses an NPI; sentences in which the environment occurs without an NPI are retained.
So for the \textit{adverbs} environment, for example, we remove sentences like ``\textit{Mary rarely ate any cookies}'' but not ``\textit{Mary rarely ate cookies}''.
For each of these nine corpora we train 3 new LMs.
Models trained on these corpora are referred to by the name \noENVName.

We run the NPI acceptability task of Experiment 2 and the DC ranking method of Experiment 3 on the nine types of \noENVName models.
A model with a robust connection between monotonicity and NPI processing should be able to learn for NPIs in the held-out environment that (i) the environment belongs to the class of environments in which NPIs are licensed, and that (ii) determining NPI acceptance should be done based on representational cues that are similar for monotonicity prediction.


\paragraph{Results}
We report the results of this experiment next to the previous results of the \fullName model.
First, we consider the NPI acceptability task, which is reported in the bottom row of Figure~\ref{fig:npi_acceptability}.
Note that each cell in this row now corresponds to a specific model type: the \textsf{ADV} result, for instance, corresponds to the accuracy of the \noENVName models in which sentences with NPIs within adverbial environments have been removed.
Our results show that the performance drops slightly for all environment classes, which can be attributed to a model's dependence on collocational cues.
However, the models are still able to adequately generalize from the other environments, in which NPIs still are encountered, to the held-out environment.
This demonstrates the semantic generalization capacities of the LM: it infers that the held-out environment in which NPIs have never been encountered shares some relevant properties with the other eight environment classes in which NPIs still occur. 

The results for the DC ranking experiment are shown in the bottom row of Figure~\ref{fig:npi_rank}.
Similar to the NPI acceptability results, the performance of the \noENVName models has dropped slightly compared to the \fullName models.
However, if the models would no longer pick up on the connection between monotonicity in the held-out environment and NPI licensing at all, these median ranks should drop to chance, i.e. around the halfway mark of the vocabulary size (25.000).
It can be seen that this is only the case for the Simple Questions environment, that was already performing poorly for the \fullName models.
Based on this we conclude that although models depend partly on collocational cues for their connection between monotonicity and NPIs, they are still able to encode a robust connection that generalizes to novel DM environments. 


%% file: sections/conclusion.tex
\section{Conclusion}\label{sec:discussion}
Based on a series of experiments, we have established the following: (1) LMs categorize environments into DM and UM; (2) LMs are overall successful with NPI licensing; (3) LMs employ similar representational cues when processing NPIs and predicting monotonicity; (4) their categories of DM and UM environments can be learned independently of NPI occurrence; and (5) their connection between monotonicity and NPI processing is robust and not solely dependent on co-occurrence heuristics.
This demonstrates that LMs have quite sophisticated knowledge of NPI licensing, which may be similar to that of humans and constitutes a vital step towards better understanding the linguistic generalization capacities of LMs.

These results raise the question: what do LMs learn about the DM and UM environments which they succeed in finding? 
Do they actually learn the inferential properties of those environments, or do they rely on some other property that DM environments have in common to categorize them as such? 
A direction for future work would be to develop methods to probe the inferential capacities of LMs and explore how they align with the DM and UM categories they construct.

Another direction for future work would be to incorporate the recent advancements on probing-based interpretability methods in our experimental pipeline \citep{hewitt2019designing, voita2020information}.
Our DC Ranking method aligns the performance of a probe with that of the language model itself, which is related to the approaches of \citet{saphra2019understanding}, \citet{DBLP:journals/tacl/ElazarRJG21}, and  \citet{lovering2021}.
Placing our methodology more firmly in this body of work will allow for stronger conclusions to be drawn regarding the semantic knowledge of current language models.

%% file: sections/appendix-npis.tex
\appendix
\section{Filtered NPIs}
\label{appendix-npis}

We here present the full list of NPIs that were used for filtering sentences from the \fullName corpus, resulting in the \noNPIName corpus. 
The method for selecting these expressions is described in Section \ref{sec:exp4}.\vspace{2mm}

\noindent\textit{A damn,
any,
any longer,
any old,
anybody,
anymore,
anyone,
anything,
anything like,
anytime soon,
anywhere,
anywhere near,
as yet,
at all,
avail,
by much,
can possibly,
could possibly,
ever,
in any,
in days,
in decades,
in minutes,
in years,
just any,
just yet,
let alone,
much help,
nor,
or anything,
set foot,
squat,
such thing,
that many,
that much,
that often,
the slightest,
whatever,
whatsoever,
yet.
}\vspace{2mm}

\noindent This resulted in a reduction of 75.062 sentences out of the 3.052.726 sentences in the original \fullName corpus (2.46\%).